\documentclass[12pt]{article}
\usepackage{graphicx}
\usepackage[round]{natbib} %comment out if you do not have the package
\usepackage{url} % not crucial - just used below for the URL 

\usepackage[utf8]{inputenc} % allow utf-8 input
\usepackage[T1]{fontenc}    % use 8-bit T1 fonts
\usepackage{hyperref}       % hyperlinks
\usepackage{url}            % simple URL typesetting
\usepackage{booktabs}       % professional-quality tables
\usepackage{amsfonts}       % blackboard math symbols
\usepackage{nicefrac}       % compact symbols for 1/2, etc.
\usepackage{microtype}      % microtypography
\usepackage{xcolor}         % colors
\usepackage{graphicx}
\usepackage{multirow} 

\usepackage{enumerate}
\usepackage{amsmath}
\usepackage{amsthm}

\newtheorem{theorem}{Theorem}

\theoremstyle{definition}

\theoremstyle{definition}
\newtheorem{example}{Example}

\newcommand{\blind}{0}

\begin{document}

\def\spacingset#1{\renewcommand{\baselinestretch}%
{#1}\small\normalsize} \spacingset{1}

%%%%%%%%%%%%%%%%%%%%%%%%%%%%%%%%%%%%%%%%%%%%%%%%%%%%%%%%%%%%%%%%%%%%%%%%%%%%%%
\if0\blind
{
  \title{\bf Privacy-aware Gaussian Process Regression}
  \author{Rui Tuo\thanks{
    The authors gratefully acknowledge NSF grants DMS-2312173 and CNS-2328395, and 2024 Early Career Collaboration Program from Texas A\&M Institute of Data Science.}\hspace{.2cm}\\
    Department of Industrial and Systems Engineering, Texas A\&M University\\
    and \\
    Haoyuan Chen\\
    Department of Industrial and Systems Engineering, Texas A\&M University\\
    and\\
    Raktim Bhattacharya \\
    Department of Aerospace Engineering, Texas A\&M University}
  \maketitle
} \fi

\if1\blind
{
  \bigskip
  \bigskip
  \bigskip
  \begin{center}
    {\LARGE\bf Privacy-aware Gaussian Process Regression}
\end{center}
  \medskip
} \fi

\bigskip
\begin{abstract}
We propose a novel theoretical and methodological framework for Gaussian process regression subject to privacy constraints. The proposed method can be used when a data owner is unwilling to share a high-fidelity supervised learning model built from their data with the public due to privacy concerns. The key idea of the proposed method is to add synthetic noise to the data until the predictive variance of the Gaussian process model reaches a prespecified privacy level. The optimal covariance matrix of the synthetic noise is formulated in terms of semi-definite programming. We also introduce the formulation of privacy-aware solutions under continuous privacy constraints using kernel-based approaches, and study their theoretical properties. The proposed method is illustrated by considering a model that tracks the trajectories of satellites and a real application on a census dataset.
\end{abstract}

\noindent%
\vfill

\newpage
\spacingset{2} % DON'T change the spacing!

\section{Introduction}
\label{sec:intro}
This work aims to present a privacy-aware supervised machine learning model. At a high level, a supervised model can be regarded as a mapping from the input space to the output space. In a traditional supervised learning paradigm, the objective is to build a supervised model from the data, with the goal of minimizing its prediction error.
With the widespread use of machine learning in various industries, businesses, public sectors, and other fields, data security and privacy protection have become pressing issues that require immediate solutions.

In this article, we discuss data privacy issues in a particular scenario. A data owner intends to release a machine learning model, built using their data, to the public. However, the owner wishes to impose limitations on the prediction accuracy of the released machine learning model, as disclosing a model with the highest possible accuracy could result in private data cost, security risks, or loss of business interests. At the same time, if the accuracy is deliberately lowered too much, the model will lose its usefulness. Hence, the owner must strike a balance between the utility and privacy of the model. 

This study is inspired by the privacy and security concerns in space situational awareness (SSA). SSA refers to the ability to view, understand, and predict the physical location of natural and artificial objects in orbit around the Earth to avoid collisions. Maintaining a timely and accurate picture of space activities becomes more critical and challenging as space becomes more congested.
%SSA is crucial for tracking and predicting the location of objects in Earth's orbit to avoid collisions, especially as space becomes more congested. 
Nations and commercial operators are working together to increase satellite operation safety by exchanging information about space objects, despite privacy and security concerns. Restrictive policies for sharing data from military-owned sensors have led to a lack of confidence in the provided data, and national security concerns are heightened due to anti-satellite weapons and other counter-space capabilities. Commercial operators also worry about competitors accessing sensitive information.

%To address these challenges, the SSA mission requires rigorous inclusion of uncertainty in the space surveillance network. Instead of high-fidelity models, which contain proprietary features that cannot be shared, data-driven surrogate models are created from simulation data and observations for different stakeholders to use in SSA tasks. The challenge lies in balancing the accuracy and privacy of these surrogate models to enable proper SSA functionality while protecting military and commercial features.

To operate under these constraints, the SSA mission needs rigorous treatment of uncertainty across the space surveillance network. Rather than exchanging high-fidelity (and often proprietary) models, stakeholders can rely on data-driven surrogate models built from simulations and observations. The central challenge is balancing accuracy and privacy so SSA functions effectively without exposing sensitive military or commercial capabilities.

A key risk is that shared estimates may let outsiders infer maneuver practices, sensing performance, or operational intent \citep{chauhan2024adversarial,ding2025privacy,liang2025privacy}. Within this setting, privacy-preserving estimation \citep{das2020privacy} offers principled control of information leakage through balancing utility against privacy via bounding state-estimation error covariance using calibrated synthetic sensor noise. Complementary work on deception detection in SSA \citep{pavur2021detecting} and surveys of secure filtering \citep{yan2024privacy,song2024nonlinear,guo2024differential} underscores how unprotected products can aid adversaries. In practice, the surrogate models take time-stamped measurements and contextual simulator states (e.g., position/velocity components) as inputs and output sensor measurements (e.g., range or angles) used by stakeholders for SSA tasks.
% While an inaccurate model can increase collisions, the broader motivation is privacy risk: shared estimates can be exploited by outside parties to infer sensitive facts such as maneuver practices, sensing capability, or operational intent \citep{chauhan2024adversarial,ding2025privacy,liang2025privacy}. Within this setting \citep{das2020privacy}, privacy-preserving estimation provides a principled control on information leakage: one formulation gives an optimization-based privacy and utility trade-off in which privacy and utility are defined in terms of lower and upper bound on the state estimation error covariance, achieved by calibrated synthetic sensor noise. Together with work on detecting deception in SSA \citep{pavur2021detecting} and surveys of secure filtering \citep{yan2024privacy,song2024nonlinear, guo2024differential}, these sources provide concrete examples beyond collision risk and show how adversaries may learn behavior or target systems if shared products are left unprotected. The surrogate models typically take as inputs time-stamped measurements together with contextual information from simulators, i.e. the state vector describing the motion of a space object such as components of position or velocity, and they output measurement vectors collected by sensors such as measuring range or angular information for a satellite.

In this work, we build Gaussian process (GP) regression \citep{cressie1993,stein1999,santner2003,RasmussenWilliams06,gramacy2020surrogates} subject to privacy constraints, with the goal of preventing the adversaries from accurately inferring the underlying true function. The key idea of the proposed method is to add synthetic noise to the data until the predictive variance of the Gaussian process model reaches a prespecified privacy level. In the proposed methodology, we formulate the optimal synthetic covariance matrix in terms of semi-definite programming. We show that the privacy-aware solutions admit explicit expressions and can be pursued efficiently.
    The proposed methodology covers privacy constraints over a continuous domain of inputs. We propose a kernel-based approach to devise an explicit formula for these solutions and study their theoretical properties. In this work, we adopt the privacy formulation by \cite{du2012privacy} to justify the proposed methodology.

The remainder of this article is organized as follows. In Section \ref{sec:review}, we introduce the notation of GP regression. The inferential privacy framework is introduced in Section \ref{sec:formulation}. Section \ref{sec:meth} presents the main proposed methodology of this work. The proposed privacy-aware GP regression is illustrated on a satellite dynamics model and a real-world application in Section \ref{sec:exp}. Concluding remarks are made in Section \ref{sec:conclusion}.

\section{Notation for GP regression}\label{sec:review}

%The remainder of this work focuses on Gaussian process regression, which can be regarded as a non-parametric extension of the Bayesian linear model (\ref{LM})-(\ref{LMprior}).

%In this study, we restrict the input and the output spaces of the mapping of our machine learning model as $\mathbb{R}^d$ and $\mathbb{R}$, respectively. To model the raw training data $(x_i,y_i),i=1,\ldots,n$, we first 

Consider a non-parametric regression $y_i=f(x_i)+e_i$, where the noise vector $e=(e_1,\ldots,e_n)^T$ follows a multivariate normal distribution $N(0,V)$ with a known covariance $V$, and $f$ is an underlying function to be estimated from the data $(x_1,y_1),\ldots,(x_n,y_n)$. In this study, we suppose that $x_i\in\mathbb{R}^d$ and $y_i\in \mathbb{R}$. We now review some basic results in the GP regression \citep{RasmussenWilliams06}. Suppose the prior for $f$ is a GP with a mean function $m(\cdot)$ and a positive definite covariance function $K(\cdot,\cdot)$. Thoughout this work, we shall use the notion $K_{A,B}$ to denote the matrix $(K(a_i,b_j))_{i j}$ given any $A=(a_1,\ldots,a_{N_A}), B=(b_1,\ldots,b_{N_B})$. Then the Bayesian posterior mean (BPM) of $f(x_*)$ at any $x_*$ is 
\begin{eqnarray}\label{GPBLP}
  \hat{f}_{\mathrm{BPM}}(x_*):=m(x_*)+K_{x_*,X}( {K}_{X,X}+ {V})^{-1}(Y-M),
\end{eqnarray}
where $Y-M:=(y_1-m(x_1),\ldots,y_n-m(x_n))^T, X:=(x_1,\ldots,x_n)$. %A nice property with $\hat{f}_{BPM}$ is that, under the GP model, it has the smallest mean squared error (MSE) among all possible predictors conditional on the data. %So in the sense of MSE, $\hat{f}_{BPM}$ is the best predictor we can ever have.

Besides the predictive capability, GP regression enables uncertainty quantification for the predictive value in terms of its predictive variance
\begin{eqnarray}\label{GPVar}
\text{Var}[f(x_*)|x_1,\ldots,x_n,y_1,\ldots,y_n]=K_{x_*,x_*}-K_{x_*,X}( {K}_{X,X}+ {V})^{-1}K_{X,x_*}.
\end{eqnarray}
With a straightforward extension and abuse of notation, (\ref{GPBLP}) and (\ref{GPVar}) hold for multiple testing points, i.e., when $x_*$ is a vector, provided that the variance in (\ref{GPVar}) is replaced by the covariance matrix.
%To present our results in this work, we will use some straightforward extensions of (\ref{GPBLP}) and (\ref{GPVar}) as follows without further notice.
%\begin{enumerate}[1)]
%    \item 
%    \item We also need the results for GP regression in which the additive error $e_i$'s are correlated. They can still be presented in terms of (\ref{GPBLP}) and (\ref{GPVar}), provided that $ {V}$ denotes the covariance matrix of $(e_1,\ldots,e_n)^T$. In this situation, the minimum MSE property of $\hat{f}_{BMP}$ remains valid.
%\end{enumerate}

In practice, the functions $m,K$, and parameters $V$ do not have to be known exactly. We can model them within certain parametric families and estimate the hyper-parameters therein from the data via standard techniques detailed in textbooks such as \cite{RasmussenWilliams06,santner2003,gramacy2020surrogates}. As the main objective of this work is not on parameter estimation, for simplicity, we assume that the model hyper-parameters are already properly estimated from the raw data so that the mean and covariance structure of the GP model is regarded as known.

%To Do: Total variance.

\section{Inferential privacy formulation}
\label{sec:formulation}
Privacy in dynamical systems is an emerging area of work and has been primarily in differential privacy \citep{cortes2016differential,dwork2011differential, dwork2014algorithmic,  Farokhi_2019, kawano2018differential,koufogiannis2017differential, mcsherry2007mechanism}, which protects the privacy of each individual record within the dataset. Methodologies have been proposed for Gaussian process regression with differential privacy guarantees \cite{smith2018differentially,honkela2021gaussian}. However, the privacy concerns with respect to surrogate models do not focus on the privacy of individual data points. Rather, it relates to the collective \textit{knowledge} represented by the entirety of the data. Such issues are referred to as \textit{inferential privacy} \citep{du2012privacy,ghosh2016inferential, song2017composition, sun2017inference}, where we are trying to bound the inferences an adversary can make based on auxiliary information. It is worth noting that the above articles mentioned multiple definitions for inferential privacy. In this work, we adopt the framework suggested by \cite{du2012privacy}, which is most suitable for GP regression.

Now we review the formulation of \cite{du2012privacy} under the current context. Under the GP regression problem stated in Section \ref{sec:review}, suppose an owner wishes to transmit the data $(X,Y)$ to a user, while protecting the true output of $f$ at a prespecified input location $s$. The owner's goal is to find and transmit an obfuscated version of $(X,Y)$, which prevents the user from accurately inferring $f(s)$ while enabling them to have access to the model output $f(x)$ at other locations with as much accuracy as possible. Here an obfuscated data is generated by a probabilistic privacy-preserving mapping of the raw data $(X,Y)$. On the other hand, the user is assumed to be passive but curious. They will try to learn $f(s)$ as much as possible based on their prior knowledge as well as the data shared by the owner. In addition, the following assumptions are made: 1) the user has the same prior distribution as the owner, which includes the complete forms of the prior mean and covariance functions of the GP model, and 2) the user has complete knowledge of the probabilistic mechanism of the privacy-preserving mapping. These two assumptions represent the \textit{worst-case} statistical side information that an adversary can have. In this work, we further assume that the user may already know $X$, so that it is only meaningful to obfuscate $Y$. We denote the obfuscated data by $(X,W)$.

Given the aforementioned background, \cite{du2012privacy} quantified the utility loss and the privacy cost as follows. Let $\mathbb{E}$ be the expectation operator with respect to the product probability measure of the GP prior and the privacy-preserving mapping; see Supplementary Materials for more technical details. The utility loss is quantified as $\mathbb{E}[d(Y,W)]$, where $d$ is some distortion metric quantifying how different $W$ is from $Y$. To introduce the privacy cost, we first consider a loss function $l(f(s),q)$, quantifying the loss of estimating $f(s)$ with some statistic $q$. Therefore, before and after the owner shares their data, the minimum expected losses are
\[c_0:=\inf\{\mathbb{E}[ l(f(s),q)] :q \text{ is a measurable function of } \Gamma\},\]
and
\[c_W:=\inf\{\mathbb{E}[l(f(s),q)|W]:q \text{ is a measurable function of } \Gamma \text{ and } W\},\]
respectively, where $\Gamma$ denotes all variables the user may have known before the data sharing, including the parameters of the GP prior and the privacy-preserving mapping, and $X$. Now define the \textit{maximum privacy cost} as $c_0-\operatorname{ess}\inf c_W$, where $\operatorname{ess}\inf$ stands for the essential infimum of a random variable.
Having the utility and privacy metrics, \cite{du2012privacy} formulated the privacy-utility tradeoff as a constrained optimization problem
\[\min c_0-\operatorname{ess}\inf c_W, \text{ subject to: } \mathbb{E}[d(Y,W)]\leq \delta,\]
for some tolerance level $\delta$,
where the minimization is taken over a set of privacy-preserving mappings.
The above setup is called a utility-aware formulation, in which the constraint is set for the utility loss and we minimize the privacy cost. In this work, we consider the privacy-aware formulation
\begin{eqnarray}\label{privacy_aware}
   \min \mathbb{E}[d(Y,W)], \text{ subject to: } c_0-c_W\leq \delta \text{ almost surely}.
\end{eqnarray}
 In view of the constraint in (\ref{privacy_aware}), the proposed methodology meets the following privacy guarantees: the maximum privacy cost (with respect to the loss function $l$) is at most $\delta$.
Solving the optimization problem (\ref{privacy_aware}) in its general form can be challenging. But we will show that for GP regression, it can be made much easier if suitable $d$ and $l$ are chosen.

%Unlike differential privacy, there is no formal definition of inferential privacy in the literature. In this work, we introduce a mathematical framework to describe and study inferential privacy. We define the goal of inferential privacy as protecting sensitive knowledge, represented by a function of the true parameter of a statistical model. The proposed framework involves the following ingredients: 1) a prior on the true parameter, which reflects the knowledge of the \textit{user}; 2) a data obfuscation scheme; and 3) a loss function to quantify the estimation error. Our definition of inferential privacy requires that no statistics can estimate the sensitive parameter at high accuracy under the joint probability measure of the prior and the sanitized data. This framework is intimately related to the classic statistical decision theory \citep{berger2013statistical}. Via the latter, it can be seen that inferential privacy is inversely proportional to the estimation uncertainty.

\section{Methodology}\label{sec:meth}
\subsection{Privacy-aware GP regression with finite sensitive input}\label{sec:ideas}

To start, we adopt the framework of \cite{du2012privacy}.
 Specifically, we consider the privacy-preserving mapping $W=Y+Z$ with $Z\sim N(0,\Sigma)$ independent of the data for some covariance matrix $\Sigma$ to be specified later. We want to choose the distortion metric $d$ and the loss function such that: 1) they have nice intuition, and 2) (\ref{privacy_aware}) can be solved easily. Our choice is $d(Y,W):=\operatorname{Tr}\big( (Y-W)^T(Y-W) \big)$, with the resulting utility loss%simply
\[\mathbb{E}\big[ \operatorname{Tr}\big( (Y-W)^T(Y-W) \big) \big]=\operatorname{Tr}\big( \mathbb{E}[ Z^T Z ] \big) =\operatorname{Tr}(\Sigma).\]
And we use the quadratic loss function $l(x,y):=(x-y)^2$. A well-known result in classical statistical decision theory \citep{berger2013statistical} states that the minimum of the expected quadratic loss function is the (posterior) variance; also see Theorem A.1 in the Supplementary Materials. Therefore, the privacy cost is $\text{Var}[f(s)]-\operatorname{ess}\inf\text{Var}[f(s)|U]$. Hence, we obtain the optimization problem
\[\min_{\Sigma}\operatorname{Tr}(\Sigma) \text{ subject to: } \text{Var}[f(s)]-\text{Var}[f(s)|U]\leq \delta \text{ almost surely}.\]
The constraint of the above problem has a direct explanation: we require the predictive variance at $f(s)$ to be no less than $\text{Var}[f(s)]-\delta$.

Use the conditional variance formula in analogy to (\ref{GPVar}), the constraint in the above problem has the expression $K_{s,X}( {K}_{X,X}+ {V}+ {\Sigma})^{-1}K_{X,s}\leq \delta$, or equivalently,
\begin{eqnarray}\label{GP_privacy}
    \delta-K_{s,X}( {K}_{X,X}+ {V}+ {\Sigma})^{-1}K_{X,s}\geq 0.
\end{eqnarray}
In addition to (\ref{GP_privacy}), we naturally require that $\Sigma$ is symmetric and positive semi-definite (PSD), denoted by $\Sigma\geq 0$, because $\Sigma$ is a covariance matrix.

In this work, we call $s$ the \textit{sensitive input} and $f(s)$ the corresponding \textit{sensitive feature}. When there are multiple sensitive inputs $s_1,\ldots,s_\gamma$, and features $f(s_1),\ldots,f(s_\gamma)$, we consider a matrix version of (\ref{GP_privacy}), given by
\begin{eqnarray}\label{constraint}
    \Delta-K_{S,X}( {K}_{X,X}+ {V}+ {\Sigma})^{-1}K_{X,S}\geq 0,
\end{eqnarray}
for some $\Delta>0$. As interpretation of (\ref{constraint}) under the framework of \cite{du2012privacy} is given in the Supplementary Materials. Putting together the utility considerations, we end up with the optimization problem
\begin{equation}\label{private_GP1}
\begin{split}
       &\min_{ {\Sigma}} \operatorname{Tr}( {\Sigma})\\
    \text{subject to: }& \Delta -K_{S,X} ( {K}_{X,X}+ {V}+ {\Sigma})^{-1}K_{X,S}\geq 0,  {\Sigma}\geq 0.
\end{split}
\end{equation}
Using the property of the Schur complement \citep{zhang2006schur} (which states that if $ {A}>0 $ and $ {C}>0$, then $ {A}- {B} {C}^{-1} {B}^T\geq 0$ if and only if $ {C}- {B}^T {A}^{-1} {B}\geq 0$), (\ref{private_GP1}) is equivalent to the semi-definite programming (SDP)
\begin{equation}\label{private_GP_single}
\begin{split}
       &\min_{ {\Sigma}} \operatorname{Tr}( {\Sigma})\\
    \text{subject to: }& 
     {\Sigma}+ {K}_{X,X}+ {V}-K_{X,S}\Delta^{-1}K_{S,X}\geq 0,  {\Sigma}\geq 0.
\end{split}
\end{equation}
%Although SDPs are polynomial-time solvable, general SDP solvers do not scale well practically. The most commonly used SDP software packages (e.g. CVX \citep{grant2014cvx} or CVXPY \citep{diamond2016cvxpy}) may struggle when the matrix dimension exceeds 1000, which is not a large sample size to construct a typical digital twin. Here we leverage the structure of (\ref{private_GP_single}) to find more efficient computational approaches. 
The solution to the above problem can be expressed explicitly. According to Theorem \ref{Th:SDP} below, (\ref{private_GP_single}) has a unique optimal point, with explicit expression 
\begin{eqnarray}\label{PSD_part}
     {\Sigma}_{opt}=(K_{X,S}\Delta^{-1}K_{S,X}- {K}_{X,X}- {V})^+,
\end{eqnarray}
where $ {A}^+$ denotes the PSD part of $ {A}$, for a symmetric matrix $ {A}$. A possible approach to compute the PSD part is to use the spectral decomposition of symmetric matrices. Given the spectral decomposition $ {A}= {O}^T\operatorname{diag}(\lambda_1,\ldots,\lambda_n) {O}$ with an orthogonal matrix $ {O}$, then $ {A}^+= {O}^T\operatorname{diag}(\lambda_1^+,\ldots,\lambda_n^+) {O}$, where $\lambda^+=\max(\lambda,0)$ for $\lambda\in\mathbb{R}$. It is important to note that $ {A}^+$ is unique (i.e., independent of the choice of $ {O}$.) 

\begin{theorem}\label{Th:SDP}
Let $ {B}$ be a symmetric matrix. Then $ {B}^+$ is the only optimal point of the minimization problem
\begin{equation}
\begin{split}\label{SDP}
    &\min_{ {A}} \operatorname{Tr}( {A})\\
    \text{subject to: }& 
     {A}-  {B}\geq 0,  {A}\geq 0.
\end{split}
\end{equation}
\end{theorem}

With a standard spectral decomposition algorithm, such as the Jacobi eigenvalue algorithm  \citep{press2007numerical}, we can pursue (\ref{PSD_part}) for problems with moderate sample sizes. The desirable synthetic noise $Z$ can then be sampled in the following steps.
\begin{enumerate}[Step 1.]
    \item  Find a spectral decomposition  $K_{X,S}\Delta^{-1}K_{S,X}-{K}_{X,X}- {V}={O}^T\operatorname{diag}(\lambda_1,\ldots,\lambda_n)  {O}$.
    \item  Sample $\omega_i$'s independently from the standard normal distribution. Compute $U=(\sqrt{\lambda_1^+}\omega_1,\ldots,\sqrt{\lambda_n^+}\omega_n)^T$. Of course, there is no need to sample $\omega_i$ if $\lambda_i\leq 0$.
    \item Compute $Z= {O}^TU$.
\end{enumerate}

Besides the Jacobi eigenvalue algorithm, 
recent developments in fast computation of the square roots of PSD matrices \citep{hale2008computing,pleiss2020fast} also enable a more scalable algorithm to find the PSD part of a sysmetric matrix, using the identity $\mathbf{A}^+=(\mathbf{A}+\sqrt{\mathbf{A}^2})/2$. 

%\begin{example}
%\subsubsection{A numerical example}
\begin{example}\label{Sec:Example_1}
We conduct a toy numerical experiment to illustrate the proposed idea and its advantages. Suppose the input points are $i/10,i=1,\ldots,9$, and the covariance function of the GP is $K(x,y)=\exp\{-10 (x-y)^2\}$. We assume that $ {V}=0$ for simplicity, i.e., there is no intrinsic random error in the raw data, which happens when, for example, the data are outputs of a deterministic computer simulation model. Suppose that we have one sensitive input $x=0.5$, with the corresponding tolerance level $\delta=0.5$, that is, we require that the predictive variance at $x=0.5$ is no less than $0.5$. The left plot of Figure \ref{fig:example} shows the values of the optimal $9\times 9$ covariance matrix $ {\Sigma}_{opt}$ in (\ref{PSD_part}) with colors: a darker color stands for a larger value. The diagonal entries of the covariance matrix give the variances of each component of the synthetic noise. It can be seen that the noise levels for $x=0.3,0.4,0.5,0.6,0.7$ are higher, which intuitively agrees with our objective of protecting $f(0.5)$. The off-diagonal entries show the correlations between components of the synthetic noise. It can be seen that the noises at $x=0.3,0.4,0.5,0.6,0.7$ exhibit strong correlations.

\begin{figure}[h]
    \includegraphics[width=\linewidth]{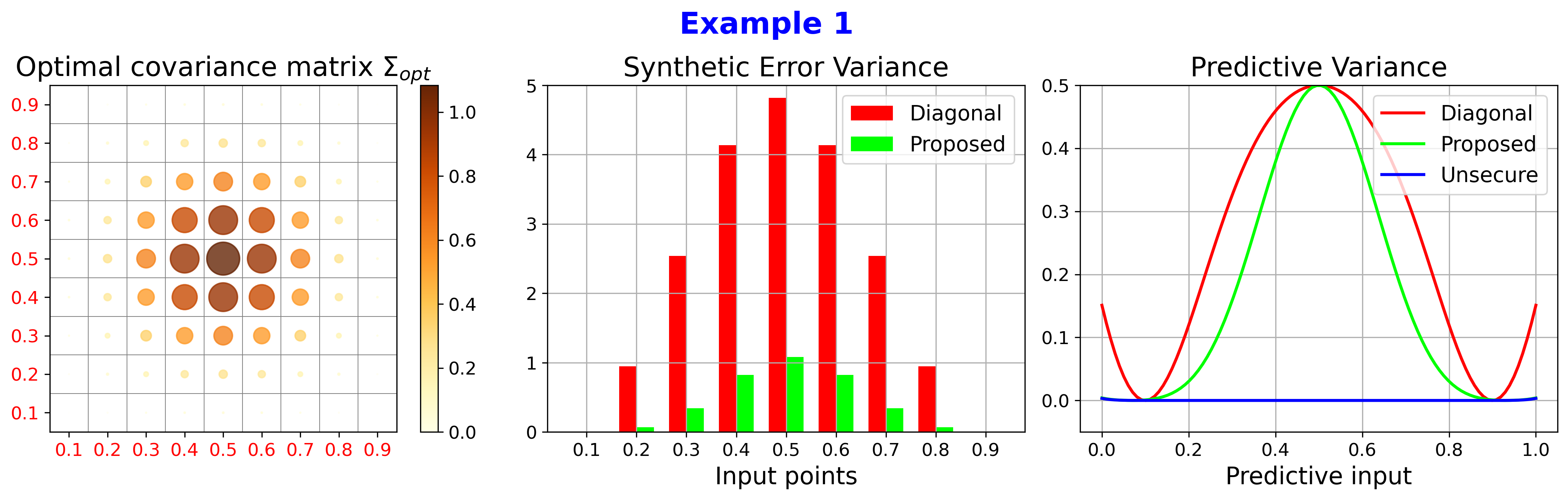}
    \caption{Numerical results for Example \ref{Sec:Example_1}. Left: Covariance matrix of the synthetic noise of the proposed method. Middle: Comparison between the synthetic noise variances of the method using independent errors (referred to as the diagonal method) and the proposed method, in order to reach the same privacy level. Right: Comparison of the predictive variance of the diagonal method, the proposed method, and the unsecured method (i.e., the ordinary GP regression).}\label{fig:example}
\end{figure}

It is worth noting that the proposed method differs from most existing data obfuscation schemes, where the synthetic error components are assumed as independent. To build privacy-aware GPs, however, using correlated error is arguably more appropriate. For comparison, we consider the problem of adding synthetic Gaussian noise whose covariance matrix is diagonal. This is simply adding one additional constraint to (\ref{private_GP_single}) to require that $ {\Sigma}$ is a diagonal matrix, which also leads to an SDP. Unlike (\ref{private_GP_single}), this SDP does not admit an explicit solution but can be pursued using an existing interior point method. As shown in the middle plot of Figure \ref{fig:example}, if we enforce the diagonal covariance matrix condition, the required variances to ensure the same privacy level are significantly inflated. The right plot of Figure \ref{fig:example} shows the predictive variances of three methods (uncorrelated synthetic noise, the proposed method, and the ordinary GP regression) with a known covariance structure. It is shown that the resulting predictive variance of the diagonal method is much larger than the proposed method, especially for $x$ near 0 or 1. This is undesirable because we only need to protect the output privacy at $x=0.5$, and adding independent noise creates an unwanted compromise in the GP model's predictive accuracy. The blue curve in the right plot of Figure \ref{fig:example} is the predictive variance of the ordinary GP regression. Its predictive variance is small, but it cannot fulfill the privacy requirement, that is, the predictive variance at $x=0.5$ should be no less than $0.5$.
\end{example}

\subsection{Kernel-based approach and infinite sensitive inputs}\label{sec:kernel}

%To define the privacy-aware solution, we need to specify the $\gamma\times\gamma$ PSD matrix $\Xi$ satisfying the constraint $K_{S,S}-\Xi>0$. A general approach is to use a positive semi-definite kernel function $H$ to define $\Xi$, i.e., let $\Xi=H_{S,S}$. In this situation, if we can ensure that $K-H$ is positive definite, we have $K_{S,S}-H_{S,S}>0$ for any $S$. Sufficient conditions for positive definite $K_{S,S}-H_{S,S}$ are given in the Supplementary Material Section C.

%Another important property of the kernel-based approach is that the synthetic noise level, given by

%is only increasing as the set of sensitive inputs $S$ increases; see Theorem \ref{Th:increase}

%\begin{theorem}\label{Th:increase}
%    Suppose $K-H$ is positive definite, and $S_1\subset S_2$ are both finite sets. Let
%    \begin{eqnarray*}
%    \Sigma_i&:=&(K_{X,S_i} (K_{{S}_i,S_i}-{H_{S_i,S_i}})^{-1}K_{S_i,X}- {K}_{X,X}-V)^+,
%    \end{eqnarray*}
%    for $i=1,2$.
%    Then $\Sigma_2-\Sigma_1\geq 0$.
%\end{theorem}

%\subsection{}

In practice, the number of sensitive inputs can be infinite. A typical example is that the user wants to protect the output values of $f$ over an open subregion of $\mathbb{R}^d$. 
%The formulation of kernel-based strongly privacy-aware GP regression in Section \ref{sec:kernel} allows for studying the limit of the solutions as the size of $S$ tends to infinity. Note that if $\Xi=H_{S,S}$, the solution to (\ref{private_GP_kernel}) is 
Recall that the optimal covariance matrix is
\begin{eqnarray}\label{kernel_solution}
    \Sigma_{opt}(S):=(K_{X,S} \Delta^{-1}K_{S,X}- {K}_{X,X}-V)^+.
\end{eqnarray}
The goal of this section is to extend the definition of $\Sigma_{opt}(S)$ to a general set $S\subset\mathbb{R}^d$. A natural idea is to consider the limit form of (\ref{kernel_solution}) with respect to a dense finite subset of the sensitive inputs. To ensure the existence of such a limit, $\Delta$ should be chosen as a suitable function of $S$.  An easy approach is to use a positive definite kernel function $H$ to define $\Delta$, i.e., let $\Delta=H_{S,S}$.

%We only need to consider the limit of $K_{X,S} (K_{S,S}-{H_{S,S}})^{-1}K_{S,X}$, as the remaining components are independent of $S$.
Under this setting, the limit does exist, and can be expressed using the theory of \textit{reproducing kernel Hilbert spaces} (RKHSs) \citep{aronszajn1950theory,paulsen2016introduction,wendland2004scattered}. 
This work concerns only RKHSs whose reproducing kernels are positive definite. The definition of these function spaces is detailed in the Supplementary Materials.
%An RKHS is a Hilbert space of real-valued functions such that each point evaluation operator is a bounded linear functional. Let $\Omega$ be a non-empty subset of $\mathbb{R}^d$. Each RKHS on $\Omega$ is paired with a unique positive semi-definite kernel function, called the \textit{reproducing kernel}. In this work, we are only interested in RKHSs with continuous and positive definite reproducing kernels. 
Denote the RKHS on $\Omega$ with reproducing kernel $\Phi$ as $\mathcal{N}_\Phi(\Omega)$, and the norm and inner product by $\|\cdot\|_{\mathcal{N}_{\Phi}(\Omega)}$ and $\langle \cdot,\cdot\rangle_{\mathcal{N}_{\Phi}(\Omega)}$, respectively. While more details of the RKHS theory are presented in the Supplementary Materials, here we need the identity
$K_{x_i,S} \Delta^{-1}K_{S,x_j}=\langle K(\cdot,x_i),K(\cdot,x_j)\rangle_{\mathcal{N}_{H}(S)}$ with any finite set $S$, which leads to the following representation of (\ref{kernel_solution}):
\begin{eqnarray}\label{kernel_continuous}
    \Sigma_{opt}(S)=\left(G(S)- {K}_{X,X}-V\right)^+,
\end{eqnarray}
where $G(S)$ denotes the matrix whose $(i,j)$-th entry is $\langle K(\cdot,x_i),K(\cdot,x_j)\rangle_{\mathcal{N}_{H}(S)}$.
Equation (\ref{kernel_continuous}) suggests a natural extension of (\ref{kernel_solution}) as $S$ can be an arbitrary non-empty subset of $\mathbb{R}^d$. 
The following Theorem \ref{Th:minimum} implies that $\Sigma_{opt}(S)$ lies in the feasible region of (\ref{kernel_solution}) under any finite subset of $S$.%, and is the smallest possible matrix satisfying this property.
%the presentation of 
%For now, we assume that $S$, the set of sensory inputs, is a general subset of $\mathbb{R}^d$. Our theory is based on 

\begin{theorem}\label{Th:minimum}
    Suppose $H$ is continuous and positive definite, and $K(\cdot,x_i)\in \mathcal{N}_{H}(S)$ for each $i=1,\ldots,n$.  The following statements are true.
    \begin{enumerate}[1)]
        \item For each non-empty $S'\subset S$, we have $K(\cdot,x_i)\in \mathcal{N}_{H}(S')$ for each $i=1,\ldots,n$, and $G(S)-G(S')\geq 0$.
        \item If a positive definite matrix $P$ satisfies $P-G(S_0)\geq 0$ for each finite subset $S_0\subset S$, then $P-G(S)\geq 0$.
    \end{enumerate}
\end{theorem}

Theorem \ref{Th:minimum} also implies that if $G(\mathbb{R}^d)$ exists, then $G(S)$ exists for each $S\subset\mathbb{R}^d$. Thus it is of interest to verify whether $K(\cdot,x_i)\in \mathcal{N}_{H}(\mathbb{R}^d)$. This is true in a trivial case when $H=\alpha K$ for $0<\alpha<1$. A more sophisticated criterion for stationary kernels is given in Supplementary Material Section D.
The covariance matrix $(G(\mathbb{R}^d)-K_{X,X}-V)^+$ ensures privacy for all input points. Thus we call it the \textit{uniformly privacy-aware solution}.

%\begin{example}
%    Under the settings of Example \ref{Ex:3}, we have 
%\end{example}

The next question is how to compute $G(S)$ numerically when $S$ is an infinite set. Theorem \ref{Th:limit} shows that, if $S$ is an infinite compact set, $G(S)$ can be well approximated by $G(S_0)$ if $S_0$ is a sufficiently dense finite subset of $S_0$.

\begin{theorem}\label{Th:limit}
Suppose $H$ is continuous and positive definite. Let $S_n$ be a sequence of finite subsets of $\mathcal{X}$. Suppose $S_n$ has a limit and denote the closure of $\lim_{n\rightarrow\infty} S_n$ as $\bar{S}$. Suppose $\bar{S}$ is a compact set.
    If $K(\cdot,x_i)\in\mathcal{N}_{H}(\bar{S})$ for each $i$, $G(S_n)\rightarrow G(\bar{S})$, and therefore, $\Sigma_{opt}(S_n)\rightarrow \Sigma_{opt}(\bar{S})$, as $n\rightarrow\infty$.
\end{theorem}

Theorem \ref{Th:limit} provides a general consistency result, but there is no rate of convergence. In the Supplementary Materials, we link this convergence rate to the approximation theory of radial basis functions in the RKHS \citep{rieger2008sampling,rieger2017sampling,rieger2010sampling,wendland2004scattered} under additional conditions. %$K,H$ are stationary and $\frac{\tilde{R}_K(\omega)}{(1+\|\omega\|^2)^\nu (\tilde{R}_K(\omega)-\tilde{R}_H(\omega))}$ is uniformly bounded for some $\nu\geq 0$.
It is worth noting that when $H=\alpha K$ for $0<\alpha<1$ and $S\supset \{x_1,\ldots,x_n\}$, $G(S)=G(\mathbb{R}^d)=\alpha^{-1}K_{X,X}$, i.e., the solution ensuring privacy at all input locations equals the uniformly privacy-aware solution.

\section{Experiments}
\label{sec:exp}
In this section, we evaluate the proposed privacy-aware GP model on a physical simulation (Sec. \ref{subsec:satellite}) and a real-world application (Sec. \ref{subsec:census}). We benchmark its performance against three methods: (1) a differentially private GP using the cloaking method (DP-Cloaking GP) \citep{smith2018differentially}, (2) a GP trained only on non-private data (Dropout), and (3) a stationary GP trained on obfuscated data generated by the privacy-aware GP to simulate a possible attack from the adversary (Stationary). For consistency, both the privacy-aware GP and DP-Cloaking GP use the same trained GP model.

\subsection{Application to tracking of space objects}\label{subsec:satellite}
%We use the data from a mathematical model that describes the trajectories of satellites to illustrate the proposed methodology.

\begin{figure}[h!]\centering
\includegraphics[width=\textwidth]{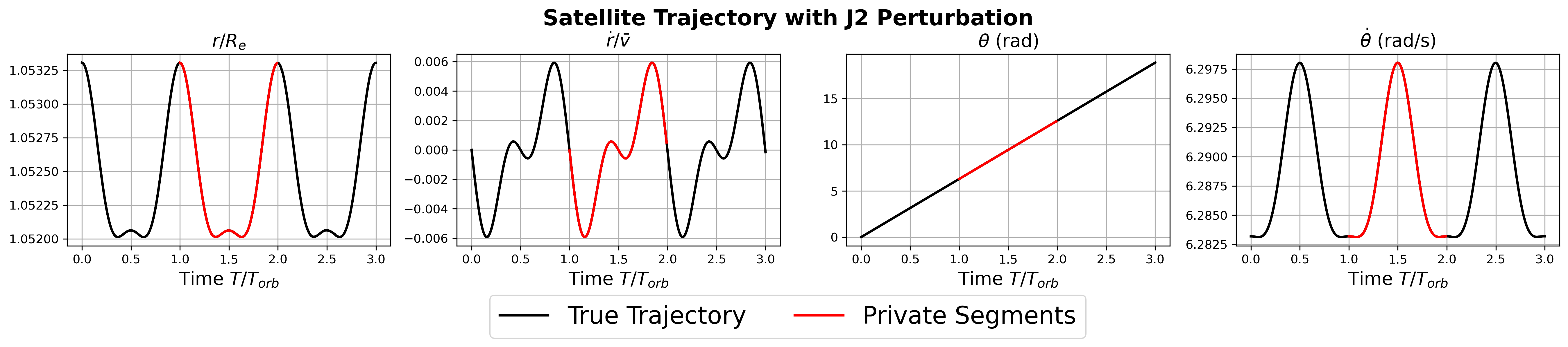}
\caption{Trajectory of the space object defined by \eqref{satDyn}. Private segments in the state trajectories are shown in red. 
%We do not want the surrogate model to provide accurate predictions in these segments.
}
\label{ltvp_nominalTraj}
\end{figure}

% \paragraph{Physics-based model for data generation}
% Here we consider a simple satellite dynamics model given by
\paragraph{Satellite dynamics model}
To illustrate the proposed methodology, we consider a simplified satellite dynamics model defined by radial and angular accelerations: 
\begin{align}\label{satDyn}
    \ddot{r} &= -\frac{\mu_e}{r^2} + \dot{\theta}^2r + \frac{3J_2}{2r^4}\left(3\sin(\theta)^2-1\right), & \ddot{\theta} &= -\frac{2\dot{\theta}\dot{r}}{r} - \frac{3J_2}{r^4}\cos(\theta)\sin(\theta).
\end{align}
% \begin{subequations}
% \begin{align}
% \ddot{r} &= -\frac{\mu_e}{r^2} + \dot{\theta}^2r + \frac{3J_2}{2r^4}\left(3\sin(\theta)^2-1\right),\\
% \ddot{\theta} &= -\frac{2\dot{\theta}\dot{r}}{r} - \frac{3J_2}{r^4}\cos(\theta)\sin(\theta).
% \end{align}
% \label{satDyn}
% \end{subequations}
% Length and time in the dynamics are normalized using $R_e$ (radius of Earth), and $T_{orb}$ (time for one orbit), respectively. 
We normalize length and time using Earth's radius $R_e$ and the orbital period $T_{orb}$, respectively. Figure \ref{ltvp_nominalTraj} depicts the satellite trajectory based on parameters listed in Table \ref{table:satData}  and the input and output variables in Table \ref{tab:input-output}, with the initial conditions:
\begin{align}
    r_0 &= \frac{R_e+h}{R_e} = 1.0533, &
\dot{r}_0 &= 0, &
\theta_0 &= 0, &
\dot{\theta}_0 &= \left(\frac{V_\theta T_{orb}}{R_e}\right)\frac{1}{r_0} = 6.2832. 
\end{align}
% \begin{subequations}
% \begin{align}
%  r_0 &= \frac{R_e+h}{R_e} = 1.0533, \\
% \dot{r}_0 &= 0, \\
% \theta_0 &= 0,\\
% \dot{\theta}_0 &= \left(\frac{V_\theta T_{orb}}{R_e}\right)\frac{1}{r_0} = 6.2832. 
% \end{align}
% \end{subequations}
% The parameters necessary to simulate the system are provided in Table \ref{table:satData}.
\vspace{-1.5em}
\begin{table}[h!]
\begin{center}
% \resizebox{\linewidth}{!}{%
\begin{tabular}{cccccc}
\toprule
$R_e$ (km) & $\mu_e$ (km$^3$/s$^2$) & $T_{orb}$ (s) & $J_2$ (km$^5$/s$^2$) & $V_\theta$ (km/s) & $h$ (km) \\
\midrule
6378.1363 & 398600.4415 & $5.48 \times 10^3$ & $1.7555\times10^{10}$ & 7.7027 & 340 \\
\bottomrule
\end{tabular}%
% }
\caption{Parameters in the satellite dynamics model.}
\label{table:satData}
\end{center}
\end{table}

\begin{table}[h!]
    \centering
    \begin{tabular}{ccc}
        \toprule
        \textbf{Variable} & \textbf{Notation} & \textbf{Description} \\
        \midrule
        Input & $T/T_{orb}$ & Time, normalized by the orbital period \\
        Output One & $r/R_e$ & Satellite's radial location, normalized by Earth's radius \\
        Output Two & $\dot{r}/\bar{v}$ & Satellite's radial velocity, normalized by $R_e/T_{orb}$\\
        Output Three & $\theta$ & Satellite's angular location\\
        Output Four & $\dot{\theta}$ & Satellite's angular velocity\\
        \bottomrule
    \end{tabular}
    \caption{Input and output variables for the proposed methodology.}
    \label{tab:input-output}
\end{table}

% \begin{table}[h!]\begin{center}\hrule\vspace{1mm}
% \begin{tabular}{ll}
% $R_e = 6378.1363$ km & $\mu_e = 398600.4415$ km$^3$/s$^2$\\
% $T_{orb} = 5.48 \times 10^3$ s & $J_2 = 1.7555\times10^{10}$ km$^5$/s$^2$\\
% $V_\theta = 7.7027$ km/s & $h$ = 340 km
% \end{tabular}\hrule\vspace{1mm}
% \caption{Parameters in the satellite dynamics model.}
% \label{table:satData}
% \end{center}
% \end{table}

% Now suppose that the goal is to create a surrogate model that can hide the trajectory information in the time interval $[1,2]$, shown in red curves in Figure \ref{ltvp_nominalTraj}.

%Consider a simple satellite dynamics model presented in the Supplementary Materials. Here we consider the input-output between the time (normalized using $T_{orb}$, the time for one orbit) and the length (normalized using $R_e$, the radius of Earth).

\begin{figure}[h!]
    \includegraphics[width=\textwidth]{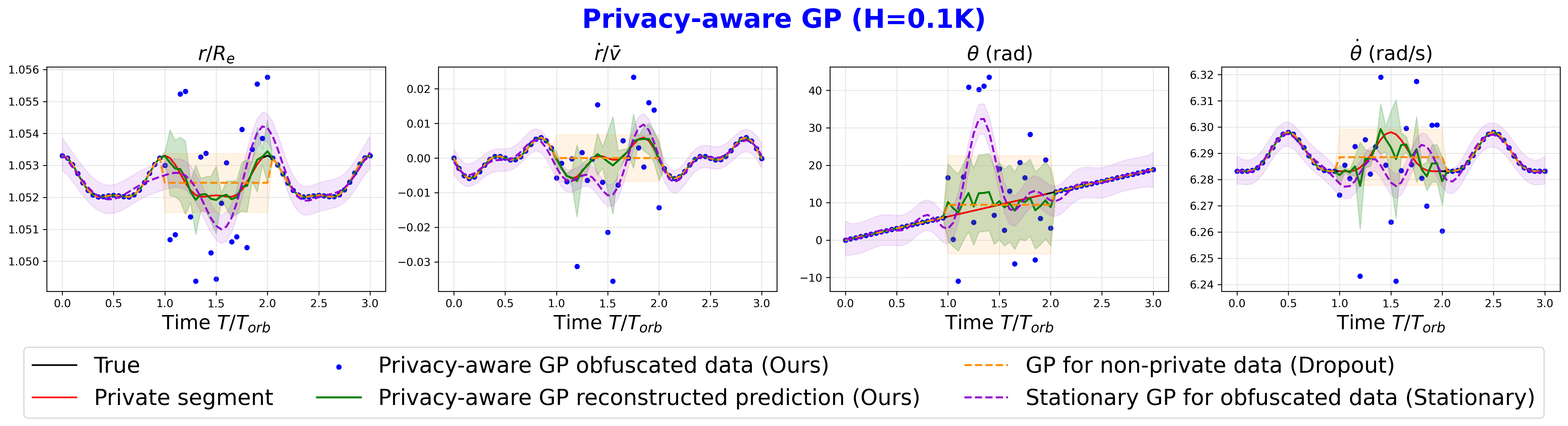}
    \includegraphics[width=\textwidth]{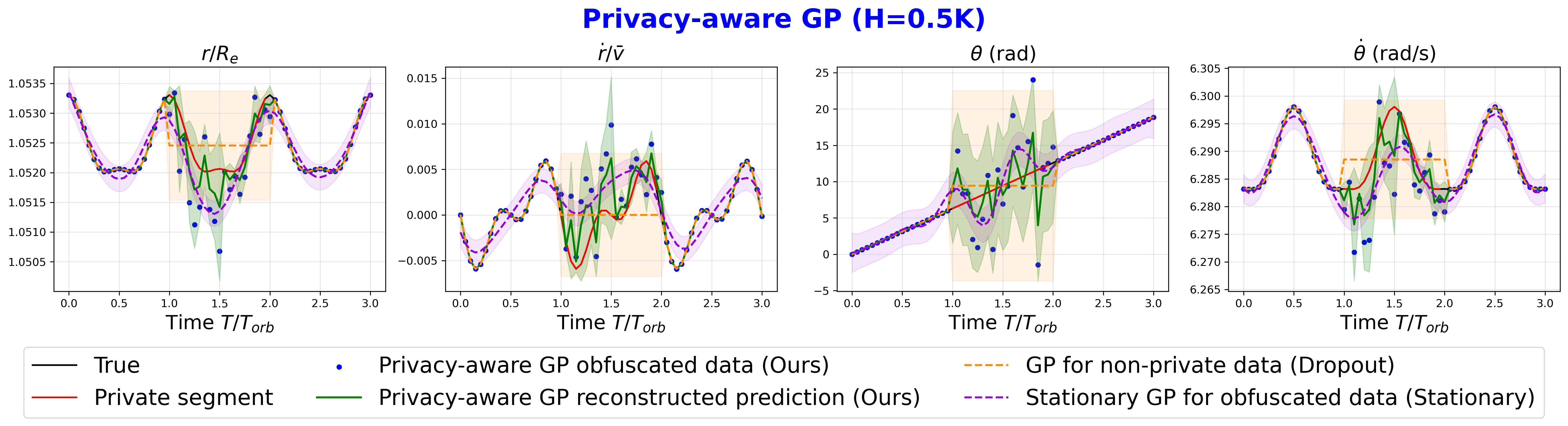}
    \includegraphics[width=\textwidth]{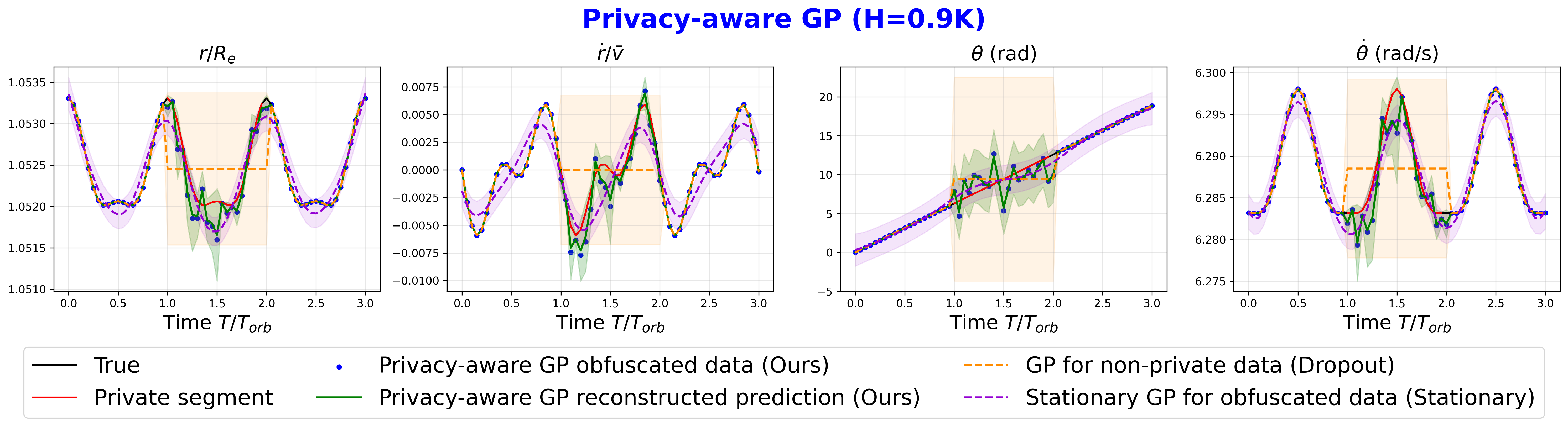}
    \caption{Privacy-aware GP regression for satellite dynamics. Shaded regions represent 95\% confidence interval of the GP based on $\mu \pm 1.96 \sigma$.}\label{fig:satellite_pa}
\end{figure}

\paragraph{Simulation}
The satellite’s location is represented in a polar coordinate system, producing four curves as shown in Figure \ref{ltvp_nominalTraj}. Our goal is to create a surrogate model that hides trajectory information within the time interval $[1, 2]$, shown in red in Figure \ref{ltvp_nominalTraj}.  For each curve, we independently train noiseless GPs with the RBF kernel $k(x,x')=\sigma^2 \exp\{ - (x-x')^2 / (2\ell) \}$ with fixed lengthscale $\ell = 1/400$. For each GP, both the training inputs $X$ and test inputs $X_*$ consist of $61$ uniformly spaced points over the interval $[0,3]$. GP hyperparameters constant mean $\beta$ and output variance $\sigma^2$ are estimated via maximum likelihood estimation. 

\begin{figure}[h!]
    \centering
    \includegraphics[width=\textwidth]{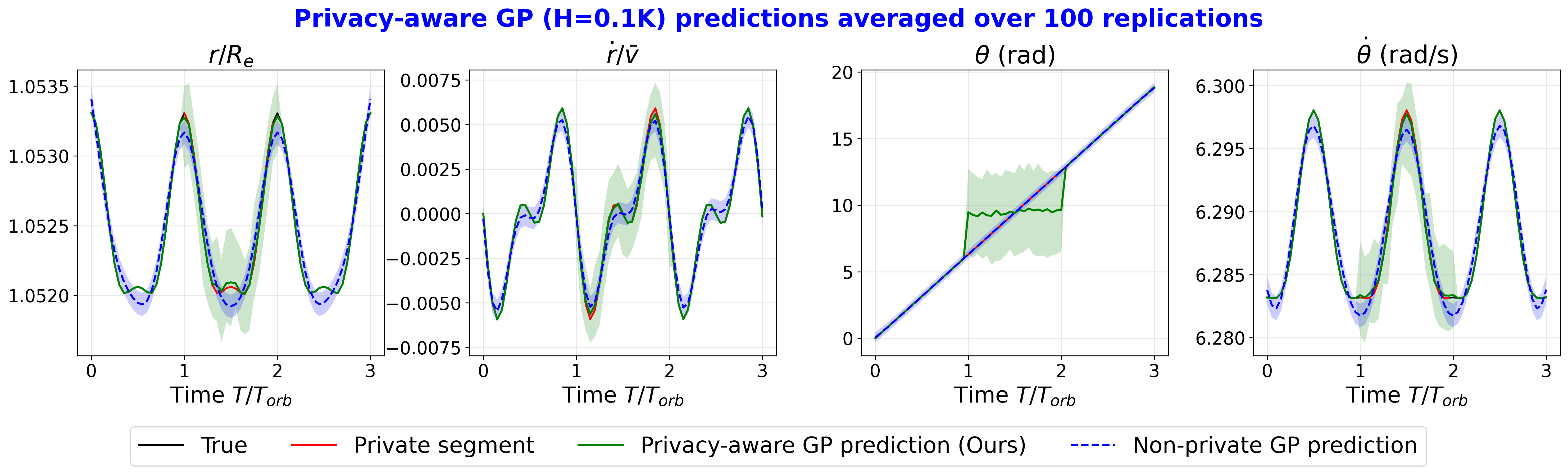}
    \includegraphics[width=\textwidth]{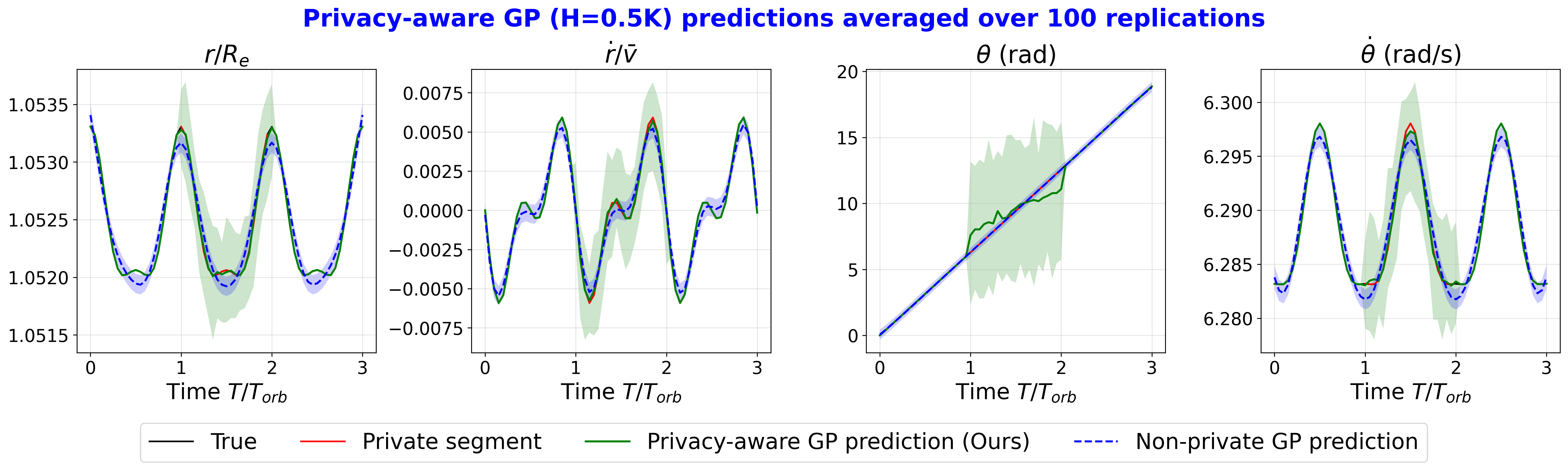}
    \includegraphics[width=\textwidth]{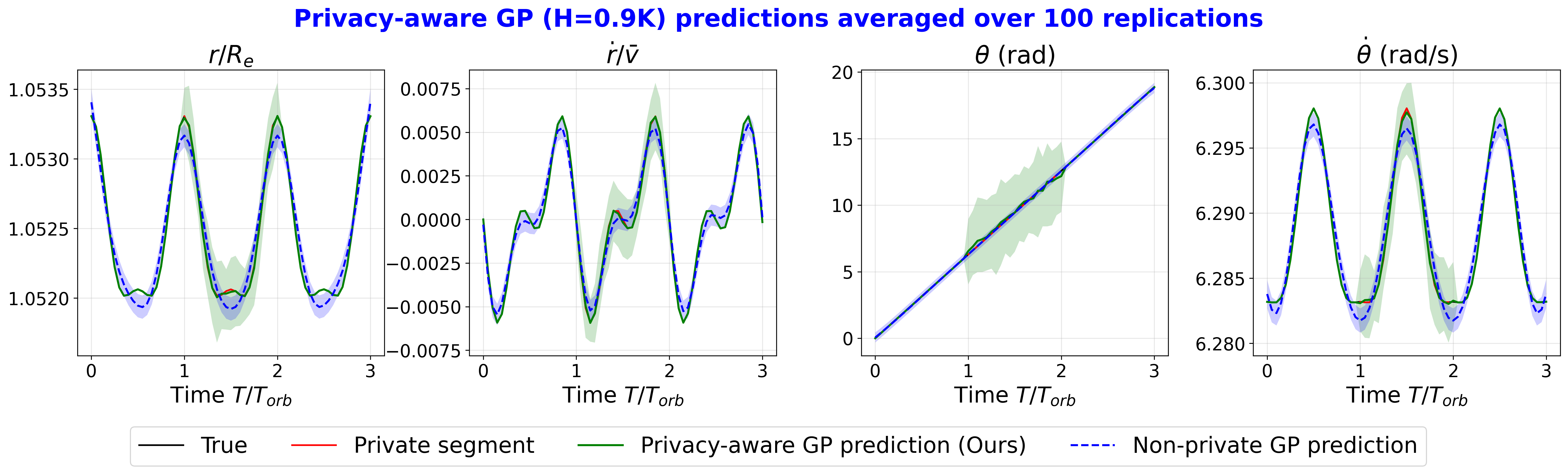}
    \caption{GP predictions for satellite dynamics averaged over 100 replications. Green shaded regions represent 96\% confidence bands of the proposed privacy-aware GP by removing the largest and smallest 2.0\% of values across 100 independent simulations. Blue shaded regions represent 95\% confidence interval of the non-private GP  based on $\mu \pm 1.96 \sigma$ where $\mu$ and $\sigma^2$ are the predictive mean and variance of the non-private GP.}
    \label{fig:satellite_pa_averaged_predict}
\end{figure}

In the privacy-aware GP, we obtain the synthetic covariance matrix $\Sigma_{opt}$ using the kernel-based approach in \eqref{kernel_solution}, with privacy region $S=[1,2]$. Following Theorem \ref{Th:limit}, we approximate $G(S)$ with $G(S_0)$, where $S_0$ consists of the densely sampled training points in the private interval $[1,2]$. We then generate obfuscated data (blue dots) and reconstruct the privacy-aware GP predictions with 95\% confidence intervals (green curves), as illustrated in Figure \ref{fig:satellite_pa}. The Dropout model, also noiseless, employs the same RBF kernel, whereas the Stationary model is a stationary GP trained directly on the obfuscated data with hyperparameters including lengthscale, noise variance, constant mean, and output variance.

Figure \ref{fig:satellite_pa} compares results for the privacy-aware GP across three privacy levels, represented by $\alpha \in \{0.1, 0.5, 0.9\}$, where the privacy constraint matrix is $H=\alpha K$. Results demonstrate that only the private segment noticeably experiences obfuscation. With $\alpha=0.1$, the privacy-aware GP predictions become highly corrupted within the private region, effectively concealing the trajectory yet maintaining predictions closer to the true data compared to the Stationary method. Conversely, with $\alpha=0.9$, the obfuscation is minimal, yielding smaller confidence intervals. The Dropout method, lacking any private data points, yields a nearly linear approximation within the private region. These results confirm that the privacy-aware GP offers consistent and robust performance.

\begin{figure}[h!]
    \centering
    \includegraphics[width=\textwidth]{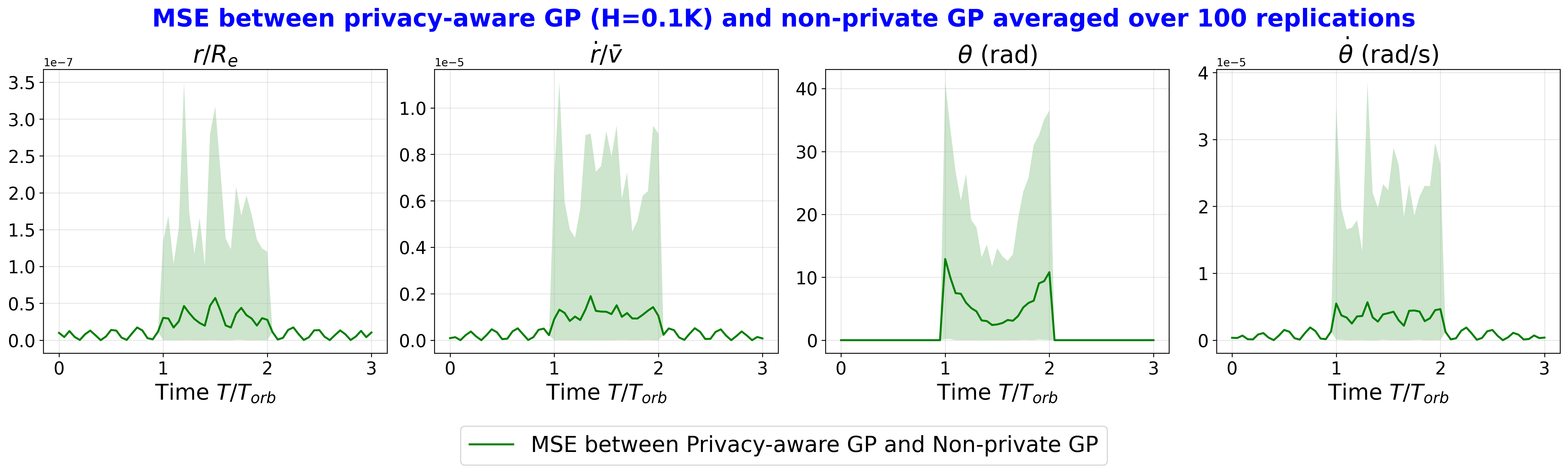}
    \includegraphics[width=\textwidth]{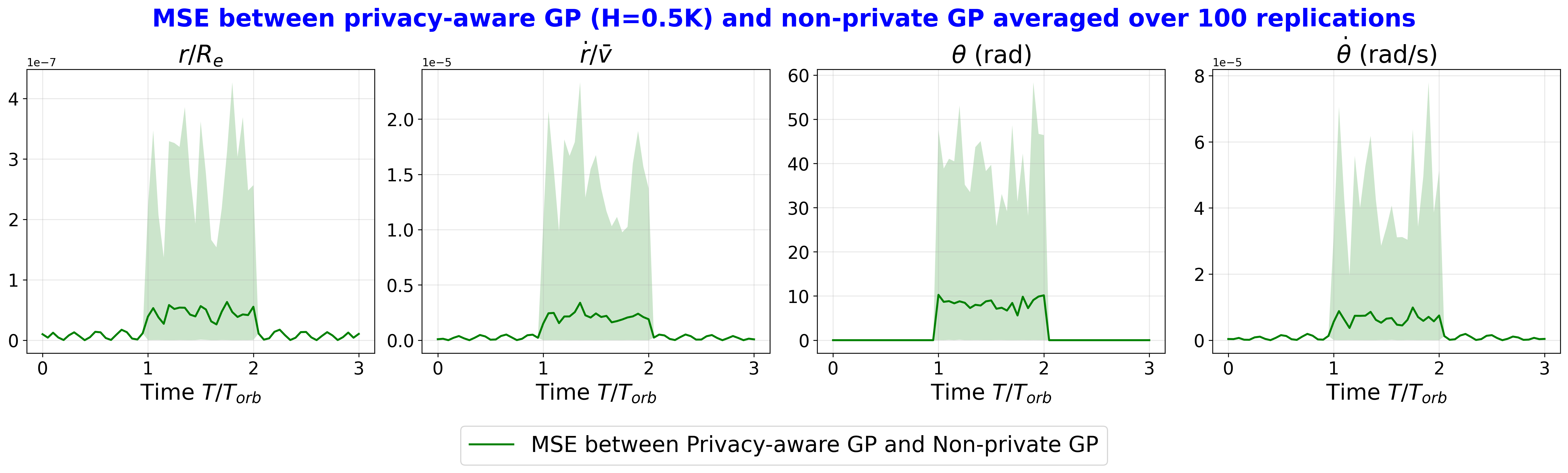}
    \includegraphics[width=\textwidth]{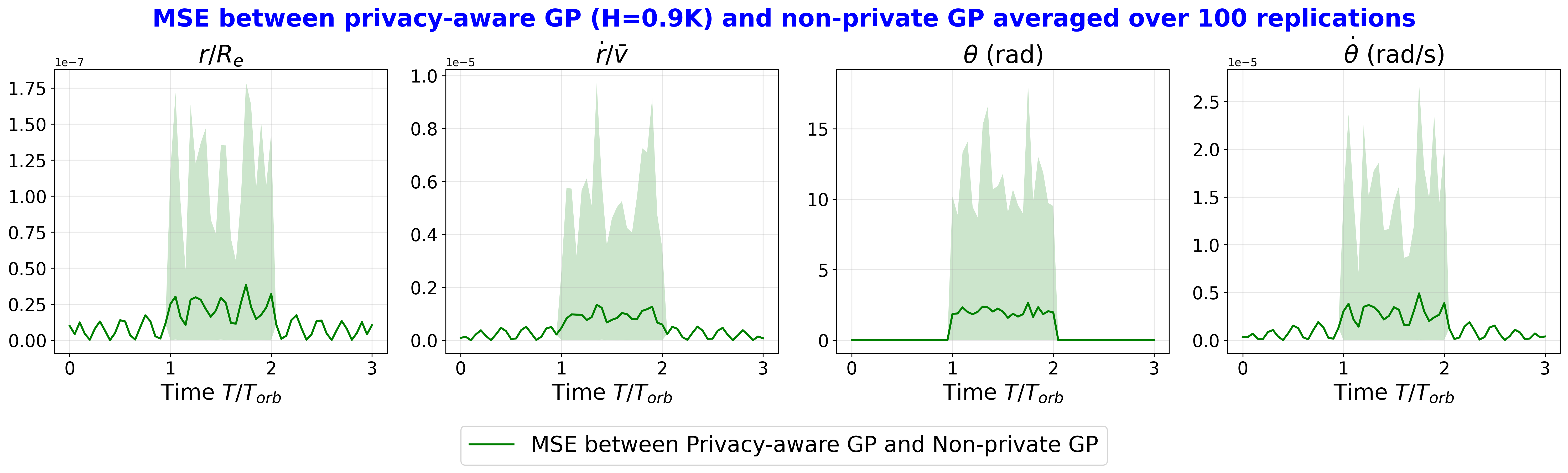}
    \caption{Mean Squared Error (MSE) between privacy-aware GP and non-private GP for satellite dynamics averaged over 100 replications. Shaded areas show 96\% confidence bands of the MSE values derived from percentile bounds (2th--98th percentiles) of 100 simulation replications.}
    \label{fig:satellite_pa_averaged_mse}
\end{figure}

Next, we examine the difference between the predictive mean of the proposed privacy-aware GP model and the non-private GP model (without data obfuscation). We investigate the distribution of $|\mu_{\text{privacy-aware}}-\mu_\text{{non-private}}|$, where $\mu_{\text{privacy-aware}}$ is the predictive mean of the privacy-aware GP and $\mu_\text{{non-private}}$ is the predictive mean of non-private GP obtained by \eqref{GPBLP}. %Because the distribution is normal, it suffices to compute the mean and the variance. %We run 100 independent simulation to estimate the mean and the variance, and use $\mu\pm2\sigma$ to build confidence bands. 
We run 100 independent simulations and %use the 96\% percentile range (removing the largest and smallest 2.0\% of values) to 
build the 96\% empirical confidence bands. The results are given in Figure \ref{fig:satellite_pa_averaged_predict}. We also measure the Mean Square Error (MSE) between the predictive mean of the proposed privacy-aware GP and non-private GP and show the results in Figure \ref{fig:satellite_pa_averaged_mse}.  It can be seen that, on the private segment $[1.0, 2.0]$, the privacy-aware GP predictive mean gets closer to both the ground truth and the non-private GP predictive mean as the privacy level $\alpha$ increases, i.e., as less data obfuscation is applied. %The results demonstrate that the predictive mean of the privacy-aware GP empirically converges to the predictive mean of the non-private GP as the privacy level $\alpha$ becomes higher.

%To evaluate the robustness of the universal error bound between the proposed privacy-aware GP model and true GP model, we average reconstructed GP predictions over multiple random seeds. For the baseline, we use the true GP predictions obtained from the GP regression equations \eqref{GPBLP} and \eqref{GPVar} applied directly to the observations without any modifications or obfuscations. Figure \ref{fig:satellite_pa_averaged_predict} and Figure \ref{fig:satellite_pa_averaged_mse} show the results for the privacy-aware GP model averaged over 100 replications. For each replication, we generate an obfuscated dataset and compute the corresponding GP predictive mean for a given seed. Averaging across 100 replications then yields (i) the average of the GP predictive means (see Figure \ref{fig:satellite_pa_averaged_predict}) and (ii) the Mean Squared Error (MSE) between the privacy-aware GP predictive mean and the true GP predictive mean (see Figure \ref{fig:satellite_pa_averaged_mse}). It's observed that the privacy-aware GP predictive means on the private segment $[1.0, 2.0]$ empirically get closer to both the ground truth and the true GP predictive means as the privacy level $\alpha$ increases, i.e., as less obfuscation is applied. The results demonstrate that the predictive mean of the privacy-aware GP empirically converges to the predictive mean of the true GP as the privacy level $\alpha$ becomes higher.

\begin{figure}[h!]
    \centering
\includegraphics[width=\linewidth]{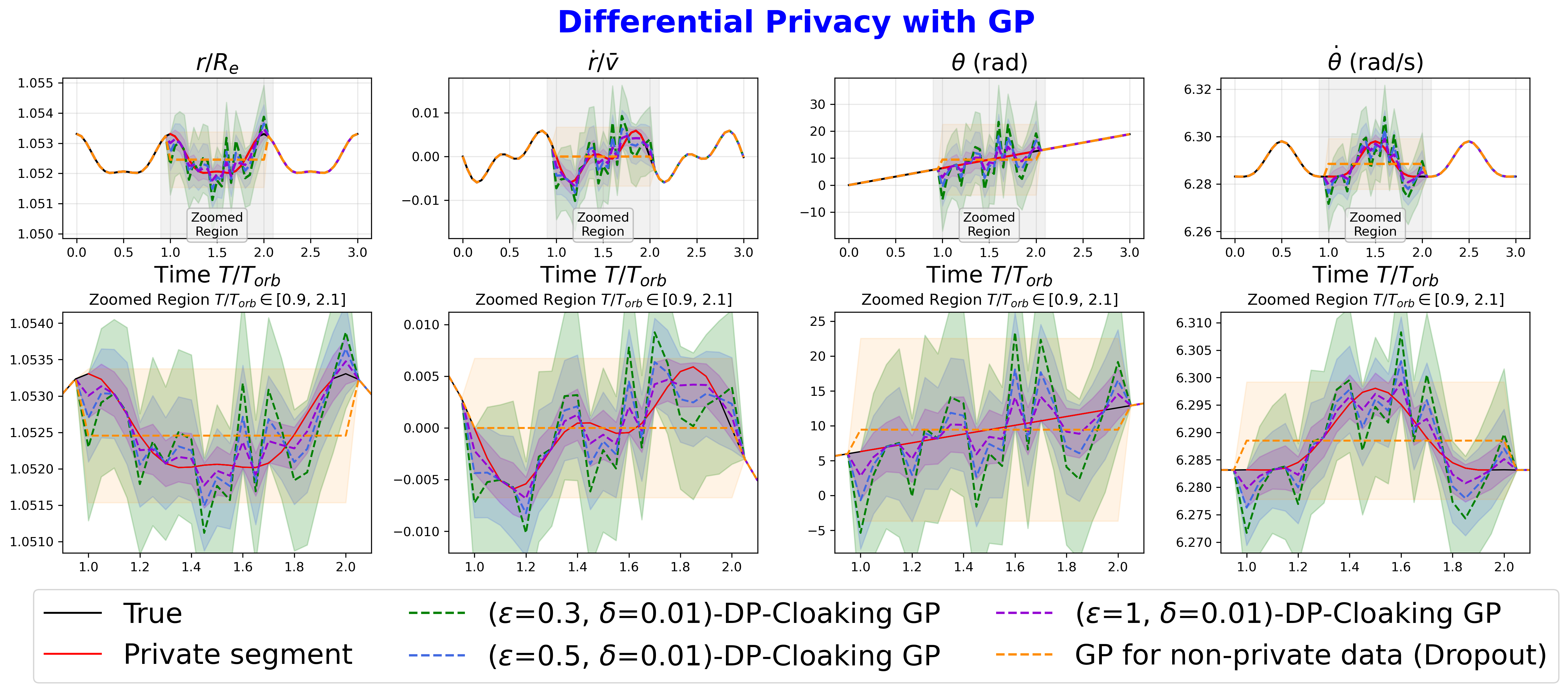}
    \caption{Differential private GP for satellite dynamics.}
    \label{fig:satellite_dp}
\end{figure}

\begin{table}[htbp]
  \centering
  \resizebox{\linewidth}{!}{%
    \begin{tabular}{l*{6}{c}}
      \toprule
      \multirow{2}{*}{\textbf{Method}} & \multicolumn{2}{c}{$\alpha = 0.1/\epsilon = 0.3$} & \multicolumn{2}{c}{$\alpha = 0.5/\epsilon = 0.5$} & \multicolumn{2}{c}{$\alpha = 0.9/\epsilon = 1.0$} \\
      \cmidrule(lr){2-3} \cmidrule(lr){4-5} \cmidrule(lr){6-7}
      & \textbf{Time (s)} & \textbf{RMSE} & \textbf{Time (s)} & \textbf{RMSE} & \textbf{Time (s)} & \textbf{RMSE} \\
      \midrule
      \textbf{$\alpha$-Privacy-aware GP} & 0.0070 $\pm$ 0.0044 & 0.3343 $\pm$ 0.5784 & 0.0072 $\pm$ 0.0031 & 0.3045 $\pm$ 0.6823 & 0.0090 $\pm$ 0.0053 & 0.1912 $\pm$ 0.3307 \\
      \textbf{Stationary GP for $\alpha$-Privacy-aware GP} & 0.0442 $\pm$ 0.0001 & 1.0412 $\pm$ 1.8002 & 0.0444 $\pm$ 0.0003 & 0.3268 $\pm$ 0.5642 & 0.0478 $\pm$ 0.0055 & 0.0663 $\pm$ 0.1133 \\
      \textbf{$(\epsilon, \delta)$-DP-Cloaking GP} & 0.4447 $\pm$ 0.0012 & 0.8878 $\pm$ 1.5342 & 0.4441 $\pm$ 0.0011 & 0.5327 $\pm$ 0.9205 & 0.4442 $\pm$ 0.0016 & 0.2663 $\pm$ 0.4603 \\
      \bottomrule
    \end{tabular}%
  }
  \caption{Performance comparison of privacy algorithms with GP for satellite dynamics.}
  \label{tab:satellite}
\end{table}

For the DP-Cloaking GP, we retain the same GP hyperparameters as in the privacy-aware GP, set sensitivity $\Delta=1$, and consider $(\epsilon, \delta=0.01)$-differential privacy with $\epsilon\in \{0.3, 0.5, 1\}$. A smaller $\epsilon$ indicates higher privacy that adds more noise to the original GP predictions. Even at $\epsilon=1$, DP-Cloaking GP introduces larger perturbations compared to our privacy-aware method, and it lacks the flexibility of adding noise directly to the training data.

Table \ref{tab:satellite} summarizes the \textit{root mean squared error} (RMSE) between the GP predictions and true values and the computational time averaged over 20 random seeds. The privacy-aware GP method achieves lower RMSE and higher computational efficiency relative to DP-Cloaking GP.

\begin{figure}[h!] 
    \includegraphics[width=\linewidth]{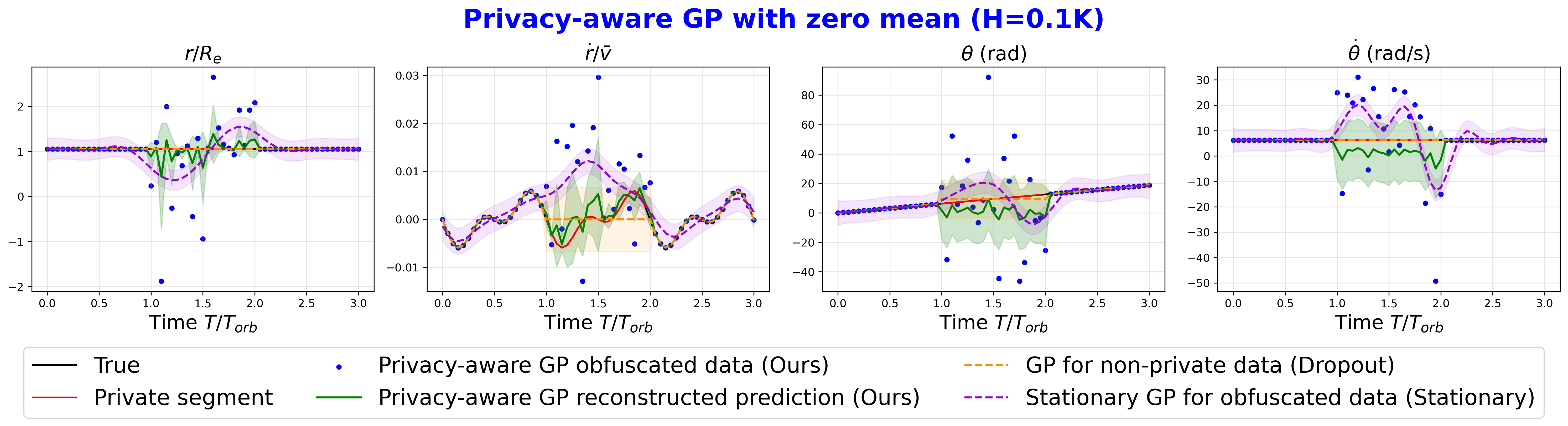}
    \includegraphics[width=\linewidth]{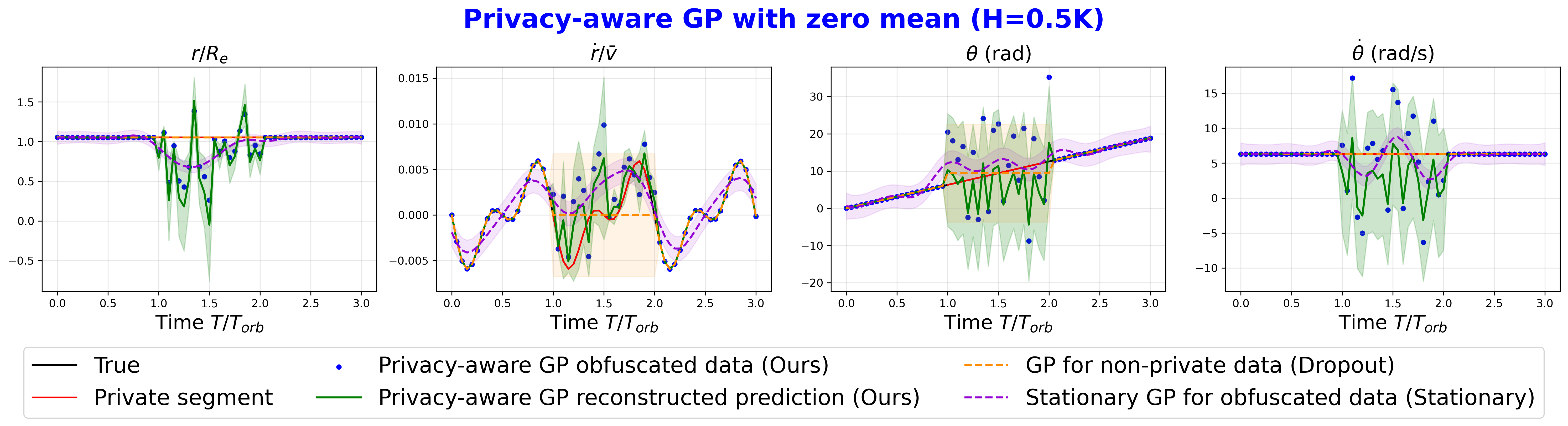}
    \includegraphics[width=\linewidth]{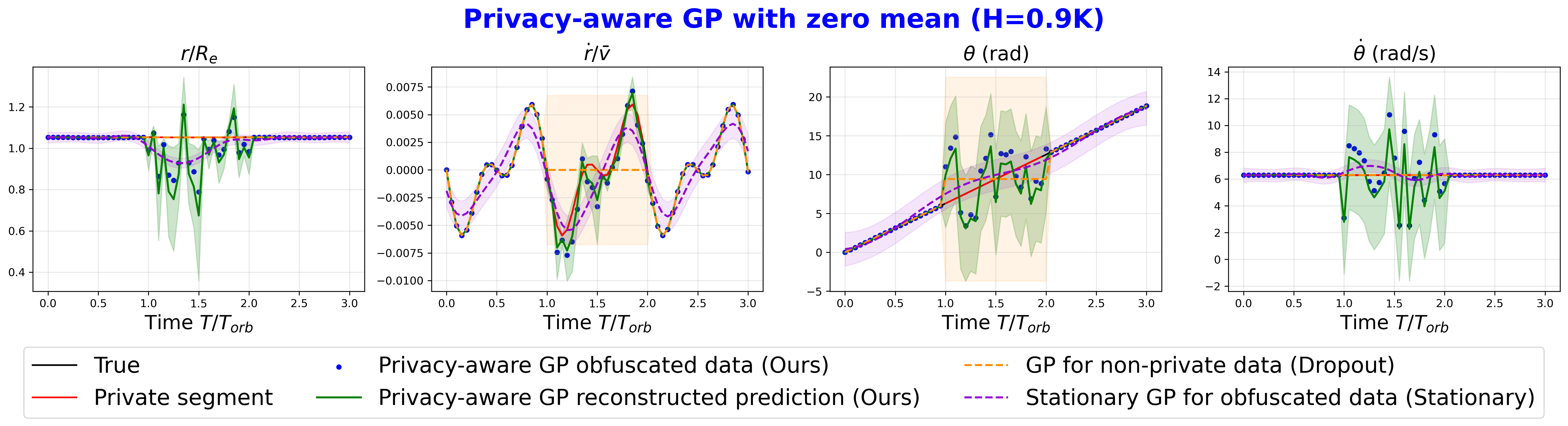}
    \caption{Privacy-aware GP regression with zero mean for satellite dynamics.}\label{fig:satellite_no_estimate}
\end{figure}

Finally, we investigate the effect of hyperparameter estimation. As shown in Figure \ref{fig:satellite_no_estimate}, setting the constant mean $\beta=0$ (i.e., no constant mean estimation) significantly increases synthetic errors, which indicates that accurate hyperparameter tuning prevents excessive obfuscation.

% Now we examine the effect of hyper-parameter estimation. For the first curve $r/R_e$, the maximum likelihood method gives $\hat{\beta}=1.052471$ and $\hat{\sigma}^2=3.854893\times 10^{-8}$. As a comparison, we consider the privacy-aware GP regression model without estimating $\beta$ and use $\beta=0$ instead. Figure \ref{fig:satellite_no_estimate} shows the result with $H=0.5K$. It can be seen that the synthetic error becomes unwantedly large. This implies that if the model hyper-parameters are not properly tuned, the proposed approach may overshoot in the synthetic noise level. 

\subsection{Application to real-world data}
\label{subsec:census}
To further demonstrate the practical utility of our privacy-aware GP approach, we apply it to real-world dataset, the person-level Public Use Microdata Sample (PUMS) dataset for Texas in 2023, provided by the U.S. Census Bureau \footnote{\url{https://www2.census.gov/programs-surveys/acs/data/pums/2023/1-Year/csv_ptx.zip}}. This dataset contains anonymized records from the American Community Survey (ACS), covering demographics, employment, education, income, and other characteristics. We consider five input features: PUMA (Public Use Microdata Area), POWPUMA (Place of Work PUMA), AGEP (Age), ANC1P (First ancestry reported code), and POBP (Place of birth), and one output target: PINCP (Total personal income). From the original data, we randomly select 1,325 records for training and 331 for testing. We define a sensitive privacy interval $S$ targeting easily identifiable or vulnerable subpopulations, specifically individuals aged 75--100 residing within PUMA areas ranging from 0 to 2000, i.e., AGEP $\in [75, 100]$ and PUMA $\in [0, 2000]$. The goal is to preserve the confidentiality of income (PINCP) information for these sensitive or potentially identifiable individuals.

\begin{figure}[h!]
    \centering
    \includegraphics[width=\linewidth]{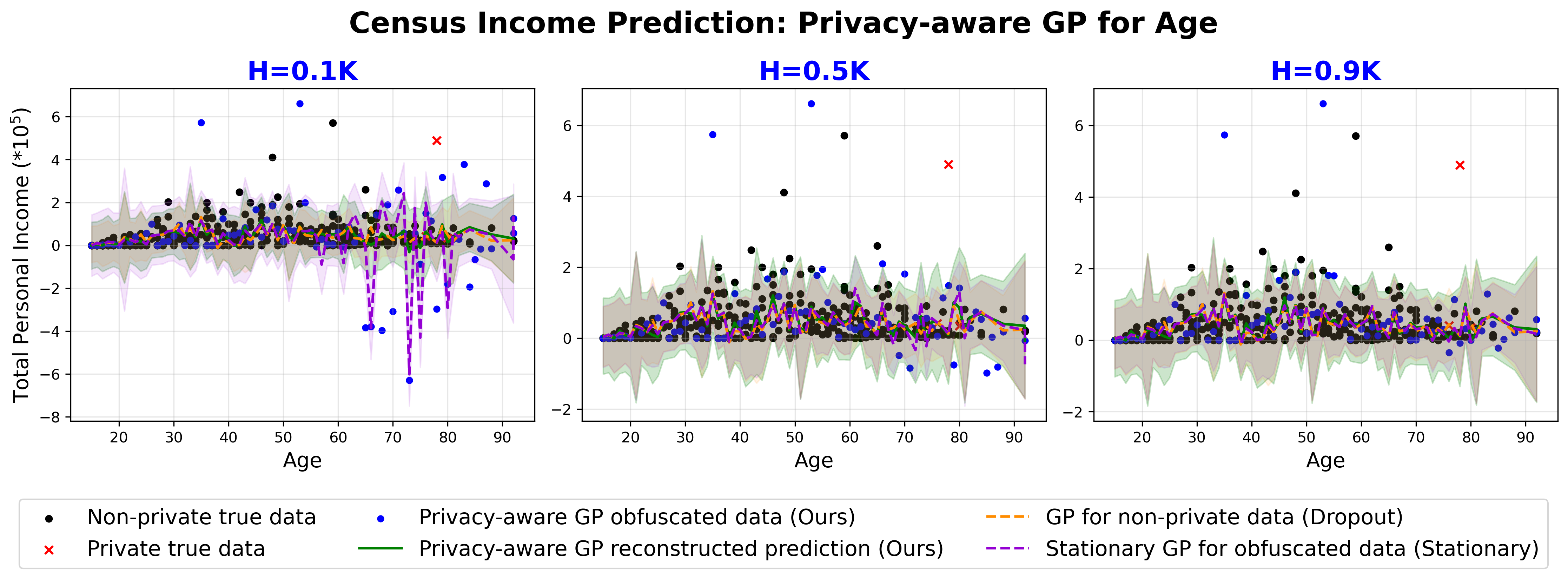}
    \includegraphics[width=\linewidth]{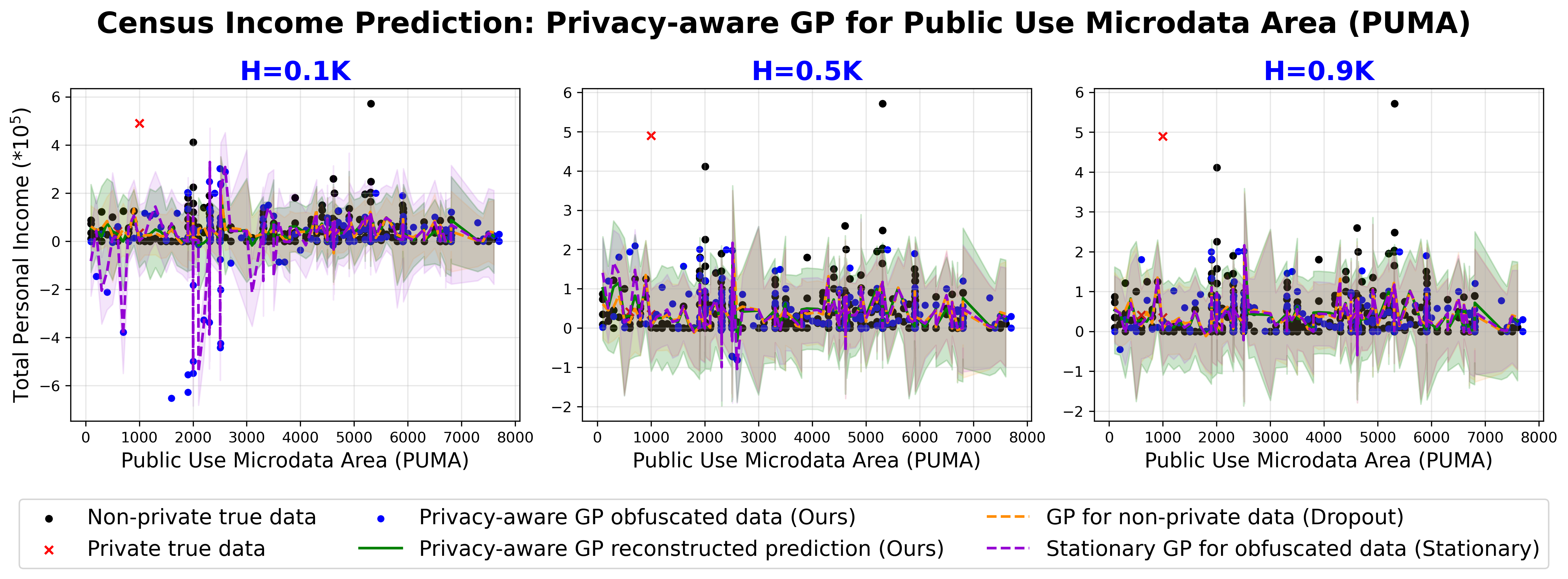}
    \caption{Privacy-aware GP regression for Census data.}
    \label{fig:census pa}
\end{figure}

\begin{figure}[h!]
    \centering
    \includegraphics[width=\linewidth]{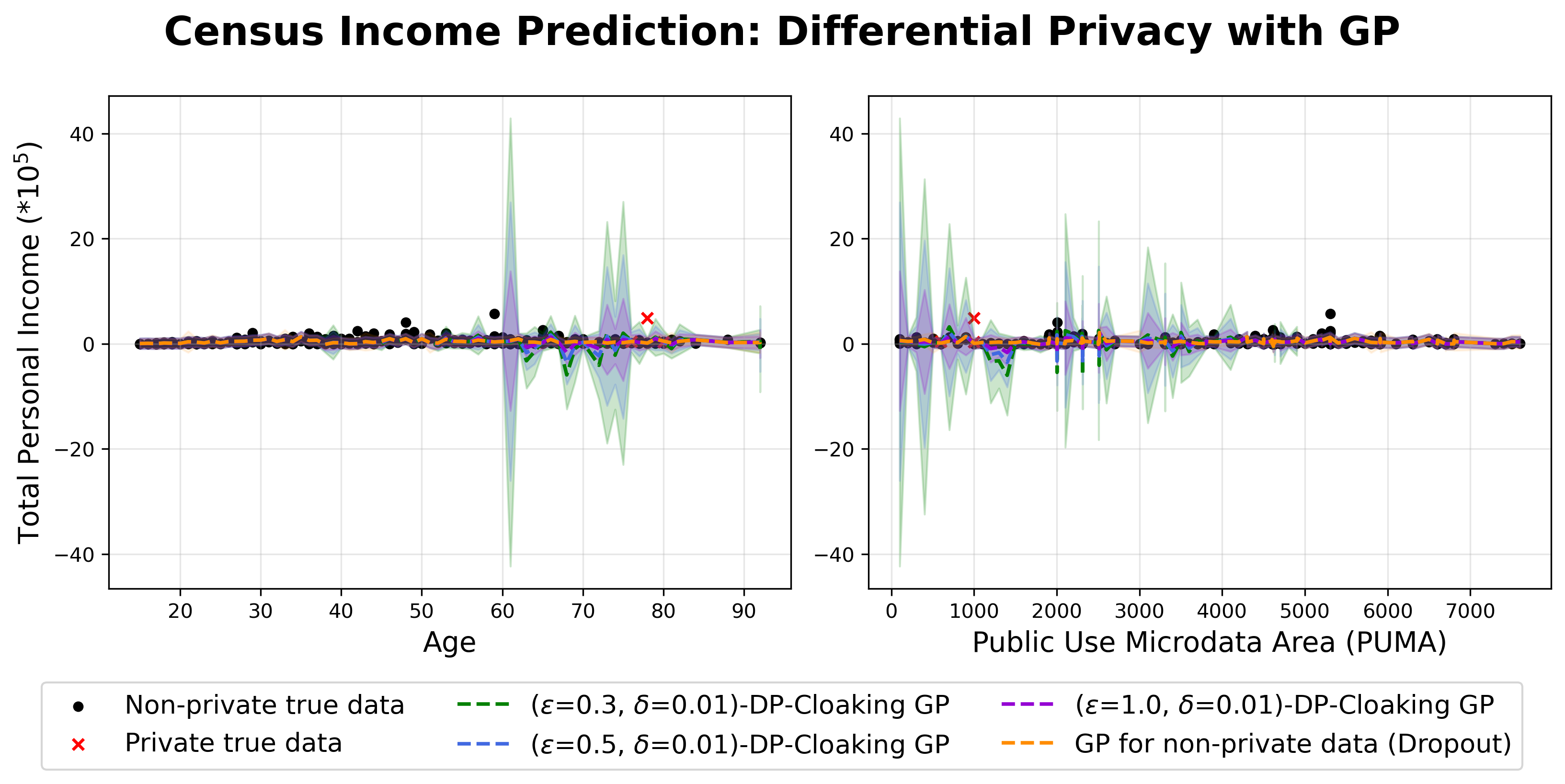}
    \caption{Differential private GP for Census data.}
    \label{fig:census dp}
\end{figure}

\begin{table}[htbp]
  \centering
  \resizebox{\linewidth}{!}{%
    \begin{tabular}{l*{6}{c}}
      \toprule
      \multirow{2}{*}{\textbf{Method}} & \multicolumn{2}{c}{$\alpha = 0.1/\epsilon = 0.3$} & \multicolumn{2}{c}{$\alpha = 0.5/\epsilon = 0.5$} & \multicolumn{2}{c}{$\alpha = 0.9/\epsilon = 1.0$} \\
      \cmidrule(lr){2-3} \cmidrule(lr){4-5} \cmidrule(lr){6-7}
      & \textbf{Time (s)} & \textbf{RMSE ($*10^5$)} & \textbf{Time (s)} & \textbf{RMSE ($*10^5$)} & \textbf{Time (s)} & \textbf{RMSE ($*10^5$)} \\
      \midrule
      \textbf{$\alpha$-Privacy-aware GP} & 2.9861 $\pm$ 0.0000 & 0.6093 $\pm$ 0.0000 & 2.9997 $\pm$ 0.0000 & 0.6048 $\pm$ 0.0000 & 2.9899 $\pm$ 0.0000 & 0.5970 $\pm$ 0.0000 \\
      \textbf{Stationary GP  for $\alpha$-Privacy-aware GP} & 1.5426 $\pm$ 0.0000 & 0.9807 $\pm$ 0.0000 & 1.5544 $\pm$ 0.0000 & 0.6389 $\pm$ 0.0000 & 1.5530 $\pm$ 0.0000 & 0.5968 $\pm$ 0.0000 \\
      \textbf{$(\epsilon, \delta)$-DP-Cloaking GP} & 25.5929 $\pm$ 0.0000 & 2.7640 $\pm$ 0.0000 & 25.6472 $\pm$ 0.0000 & 1.7148 $\pm$ 0.0000 & 25.5352 $\pm$ 0.0000 & 1.0178 $\pm$ 0.0000 \\
      \bottomrule
    \end{tabular}%
  }
  \caption{Performance comparison of privacy algorithms with GP for census data.}
  \label{tab:census}
\end{table}

We first train a GP model using the RBF kernel on the training samples, where hyperparameters (constant mean $\beta$, output variance $\sigma^2$, lengthscale $\ell$, and noise variance $\sigma_{\text{noise}}^2$) are optimized by maximizing the log-likelihood via Adam optimizer over 100 epochs. Next, we apply this trained GP model to both the proposed privacy-aware GP model and the differential privacy GP model, each tested under various privacy levels. Figures \ref{fig:census pa} and \ref{fig:census dp} illustrate predicted income (PINCP) as a function of the sensitive features age (AGEP) and location (PUMA) for privacy-aware and differentially private GPs, respectively.

Similarly to Sec. \ref{subsec:satellite}, we compare our proposed privacy-aware GP method against two baseline models: (1) Dropout, where training data within $S$ are removed, and (2) Stationary, where obfuscated data obtained from privacy-aware GP model are used for a stationary GP (see Figure \ref{fig:census pa}). Table \ref{tab:census} summarizes the RMSE between the reconstructed GP predictions and the true values, along with the computational time, averaged over 20 random seeds that generate synthetic noise in each privacy algorithm. Results demonstrate that our proposed privacy-aware GP consistently achieves lower RMSE values compared to the DP-Cloaking GP method and requires significantly less computational time. As a result, our approach provides a simpler, more efficient, and more effective solution for privacy preservation in practical real-world scenarios.

\section{Discussion}\label{sec:conclusion}

In this work, we establish a new theoretical and methodological framework for privacy-aware GP regression. Like any data science technique that addresses privacy concerns, the proposed method safeguards privacy at the cost of estimation and predictive accuracy. While the privacy levels are fully in control, the proposed method does not guarantee fit-for-use predictive outputs. Practitioners should validate the utility of the resulting privacy-aware GP models by checking whether the predictive confidence bars are acceptable. %Failure to do so may result in a \textbf{negative societal impact} of presenting unsatisfactory and misleading predictions.
%Whether the weakly privacy-aware problem (\ref{private_GP_weak}) has a unique solution is an unaddressed theoretical question and requires further investigation.
Similar to the usual GP regression, finding the synthetic covariance matrix faces computational challenges when the sample size is large. Some recent techniques in scalable GP regression %\citep{chen2022kernel,ding2021sparse,gramacy2015local,grigorievskiy2017parallelizable,hartikainen2010kalman,hensman2013gaussian,katzfuss2021general,nickisch2018state,plumlee2014fast,quinonero2005unifying,rahimi2007random,titsias2009variational,wang2019exact,wilson2015kernel, yadav2022kernel} 
can be adapted to compute certain components in the proposed method, such as $G(S)$, efficiently.

%In Section \ref{subsec:satellite}, we only consider ordinary kriging, i.e., using stationary GPs as the prior.
%As an illustrative example, here we do not consider a physics-informed GP regression. We use an ordinary kriging instead.

\bigskip
\begin{center}
{\large\bf SUPPLEMENTARY MATERIAL}
\end{center}

\begin{description}

%\item[Supplementary Materials for ``Privacy-aware Gaussian Process Regression'':]

\item[PDF Supplement:] This document contains related and extension results of the main article, including a rigorous framework for inferential privacy, technical proofs, relevant RKHS theory and criteria for positive definite kernels, and additional convergence analysis for Theorem \ref{Th:limit}. (PDF file)

\item[Code Repository:] This ZIP archive contains Python implementations of the proposed methods and the experimental pipelines reported in the main article, enabling full reproducibility of all results. The code repository is also available at \url{https://github.com/hchen19/privacygp}. (ZIP file)

\end{description}

\begin{center}
{\large\bf ACKNOWLEDGEMENTS}
\end{center}

The authors are grateful to the Editor, the Associate Editor, and two referees for their very helpful comments and suggestions.

\begin{center}
{\large\bf DISCLOSURE STATEMENT}
\end{center}

The authors report that there are no competing interests to declare.

\bibliography{references}

\end{document}